# Artificially Fluent:

# Swahili AI Performance Benchmarks Between English-Trained and Natively-Trained Datasets

**Sophie Jaffer***, Simeon Sayer

Phillips Exeter Academy, Harvard University

sophie.jaffer@gmail.com, ssayer@fas.harvard.edu

*Primary Author

**Abstract**

As large language models (LLMs) expand multilingual capabilities, questions remain about the equity of their performance across languages. While many communities stand to benefit from AI systems, the dominance of English in training data risks disadvantaging non-English speakers. To test the hypothesis that such data disparities may affect model performance, this study compares two monolingual BERT models: one trained and tested entirely on Swahili data, and another on comparable English news data. To simulate how multilingual LLMs process non-English queries through internal translation and abstraction, we translated the Swahili news data into English and evaluated it using the English-trained model. This approach tests the hypothesis by evaluating whether translating Swahili inputs for evaluation on an English model yields better or worse performance compared to training and testing a model entirely in Swahili, thus isolating the effect of language consistency versus cross-lingual abstraction. The results prove that, despite high-quality translation, the native Swahili-trained model performed better than the Swahili-to-English translated model, producing nearly four times fewer errors: 0.36% vs. 1.47% respectively. This gap suggests that translation alone does not bridge representational differences between languages and that models trained in one language may struggle to accurately interpret translated inputs due to imperfect internal knowledge representation, suggesting that native-language training remains important for reliable outcomes. In educational and informational contexts, even small performance gaps may compound inequality. Future research should focus on addressing broader dataset development for underrepresented languages and renewed attention to multilingual model evaluation, ensuring the reinforcing effect of global AI deployment on existing digital divides is reduced.

## II. Introduction

While the Internet connects more of humanity than any other technology, the language and formats it connects us in are not always representative of the billions of humans who use it. About 50% of the Internet is in English, as are many of the frameworks that enable knowledge distribution [1]. Interestingly, only about 19% of the world actually speaks English [2], making English disproportionately common online. In contrast, less than 0.1% of the Internet is in Swahili, but about 1.2% of the world speaks the language [1, 3]. The discrepancy between the language's use on the Internet versus in person creates a difference of over 10 times. This issue is compounded by the advent of LLMs, which heavily use the internet as a data source [4]. So, while the potential of language models grows, it does so disproportionately to serve those who are already online, usually English speakers. Speakers of minority languages may benefit from the affordances of LLMs, such as stronger education and as an assistant in commerce, but may be left behind because their languages are not as represented in these models.

   This paper aims to highlight the risks surrounding this inequity by examining relative accuracies between mono and multilingual models. Monolingual models describe LLMs that have been trained and are used in one language, usually English. Multilingual models describe LLMs that have been trained in multiple languages and can translate between them as such. In this experiment, a monolingual Swahili model is tested against a SOTA English model to classify Swahili news data into three categories: business, entertainment, and sport.

   Additionally, the experiment aims to investigate the role of knowledge representation within these models. There are many successful efforts that have allowed for multilingual models - the latest generation of OpenAI, Llama and Claude models can internally perform translations between languages, and multilingual reasoning. What is not clear, and what this paper aims to explore, is if these abilities are truly equitable between languages. For instance, it is not proven that a teenager asking ChatGPT for help with their math homework in their native Senegalese is helped as efficiently, or even uses the same knowledge, as their British counterparts. True language equitability is hard to define objectively, but would be evidenced by the language of choice having no effect on the quality of the responses provided by the model, as evaluated on any given metric.

Indeed, in many parts of the world, the effects of Anglo-centerism mean that there is an increased pressure to learn the English language in order to interact with the largest journals, engage expert colleagues and to understand cutting edge developments. This paper creates an experiment to analyze this broader issue of how multilingual models compare to monolingual models, by training and testing exclusively in Swahili in the context of traditional classification experiments.

**III. Background and Datasets**

There have been many efforts to understand the role of language on knowledge representation within large language models, and to extend natural language processing techniques to all forms of written language. The ability to detect and parse language for this purpose has too been an area of research. For instance, early questions of how to apply the word splitting techniques to languages that do not have a space between words have shaped the development of natural language processing [5].

In the recent generation of breakthroughs in language based machine learning, similar multilingual research has taken place. With the dawn of the BERT model system [6], presented by a Google AI Language team in 2018, efforts to extend and train new multilingual versions of this technology have been widespread. In 2022, SwahBERT [7] was trained and released by a Korean based team that aimed to offer BERT model structures and weights specifically designed for Swahili.

While many of these models are shared with performance review in mind, it is not often discussed within literature why there may be performance differences between these language models, and whether or not it is optimal to do so in the face of new, inherently "multilingual" models.

Language-specific datasets are not universally available. While it may be possible to find detailed lectures and textbooks on some subjects in French, it is not necessarily true that the same density and quality of data exists in Filipino. This data inequity means running wide-reaching experimentation is difficult.

This is also true for Swahili, where there are not many large datasets, and certainly not many that are organized for a specific classification task. As such, this paper uses one of the readily available datasets that offers this, in the form of news classification of articles. Originally released by Davis [8], the dataset includes news articles and their corresponding categories, and was generated through support given by the "Knowledge For All" foundation, which is supported by UNESCO to bring technological innovation to underrepresented regions and languages. It was used as a competition dataset in 2020 through the Zindi project, which aims to foster AI development and adoption in underrepresented countries [9].

To compare the abilities and results, an English language equivalent dataset sourced from the BBC was used [10], originally collected and categorized for a document-clustering study, but conveniently used the same news categories as the Swahili model [11].

**IV. Methodology/Models**

In order to examine the differences between how language is perceived by BERT-style transformers, and to test the hypothesis, a comparison has to be drawn between two languages and their corresponding models. For the purposes of this experiment, Swahili and English were selected.

Swahili was selected due to one of the authors having a Tanzanian family that speaks Swahili. Additionally, Swahili is a primarily oral language that has only been documented recently. Still, there are very few resources online to train and test Swahili datasets on. More recently, however, companies like Mozilla have invested more resources into documenting the Swahili language for the purpose of developing and updating LLMs [12]. English was selected as a comparative language due to the wide variety and amount of resources for training LLMs. Indeed, nearly all generative AI models have been trained in English and are based on English data.

To expand on the initial hypothesis, this project aimed to investigate whether or not keeping data in its native language for both training and testing yielded better results than training in one language, and testing on an auto-translated version of another. In short, this project aims to explore whether translating Swahili into English to test on a model that was trained in English is better or worse than simply training the model in Swahili in the first place, and keeping all training and testing results in the same language.

Therefore, by using the news classification task as a benchmark, two pre-trained transformers, SwahBERT for Swahili [7] and Google's monolingual BERT model for English [13] were trained and tested on the corresponding news classification datasets. Using the English news data, BERT was retrained to classify any given text to be one of the unique news categories - Business, Sport, or Entertainment. Using the Swahili news data, SwahBERT was similarly trained.

While the relative performance of these two categories is interesting, it does not allow for a direct comparison because the language or the nature of the datasets could be different. For instance, perhaps Swahili sports stories have a very recognizable structure or vocabulary that allows the model to more easily detect it, than in the English dataset. A better comparison that explores how data might actually be used in the context of monolingual models is to examine whether or not translating the Swahili data into English and testing it on an English trained model is comparable to using that same test set on the original Swahili model. Figure 1 demonstrates this testing structure.

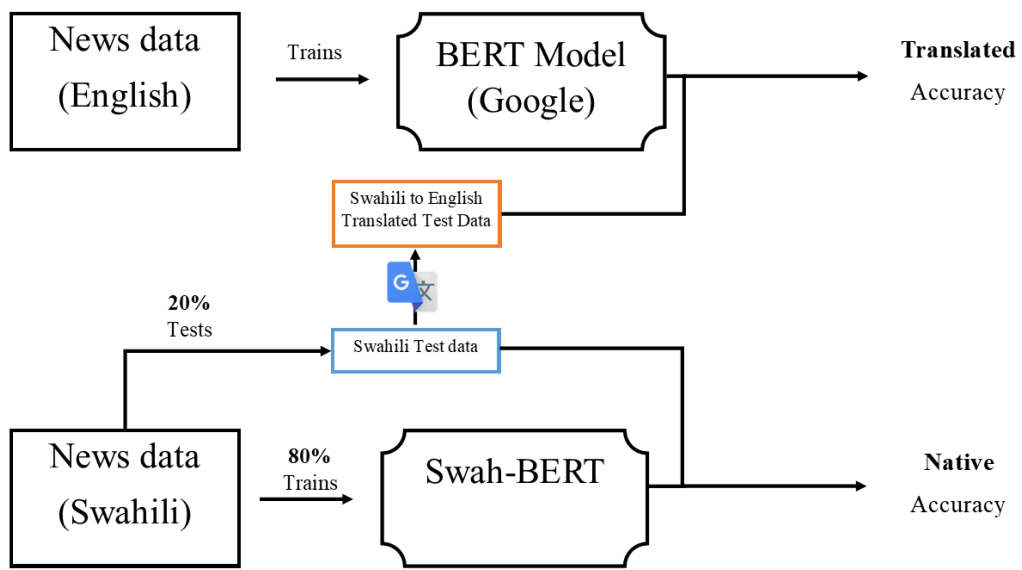

*Fig 1*. Training Structure Diagram

Alongside the two different pretrained BERT models, the final method that is relied upon to perform this experiment is a translation engine. This project used the Google Translate Python API [https://pypi.org/project/googletrans/] to automatically convert the Swahili test set into English, so that its performance could be measured on the English model. The models were accessed via Hugging Face and were run in a Google Collaboratory notebook.

**V. Results and Discussion**

The results of running the above methodology yielded the following results:

| Language Group | Accuracy | Error Rate | Error Ratio (Eng:Swah) |
| --- | --- | --- | --- |
| Swahili model on Swahili data | **99.64%** | **0.36%** | 1 |
| English model on Swahili to English translated data | 98.53% | 1.47% | 4.083 |

*Fig 2*. Accuracy Comparison

This appears to demonstrate that training a model in its native language provides higher accuracy than translating that data and testing it on a different language. The exact nature of the errors that the model made are represented in the confusion matrices below, although in both cases the error was relatively small, and the errors were evenly distributed. This demonstrates that the translation process did not introduce systematic errors, such as routinely confusing sport and entertainment, for example, but rather a general increased likelihood of misunderstanding. This indicates that the additional translation will always add an additional layer of error to the model, regardless of topic.

| Language Group | Confusion Matrix |
|---|---|
| Swahili model on Swahili data | 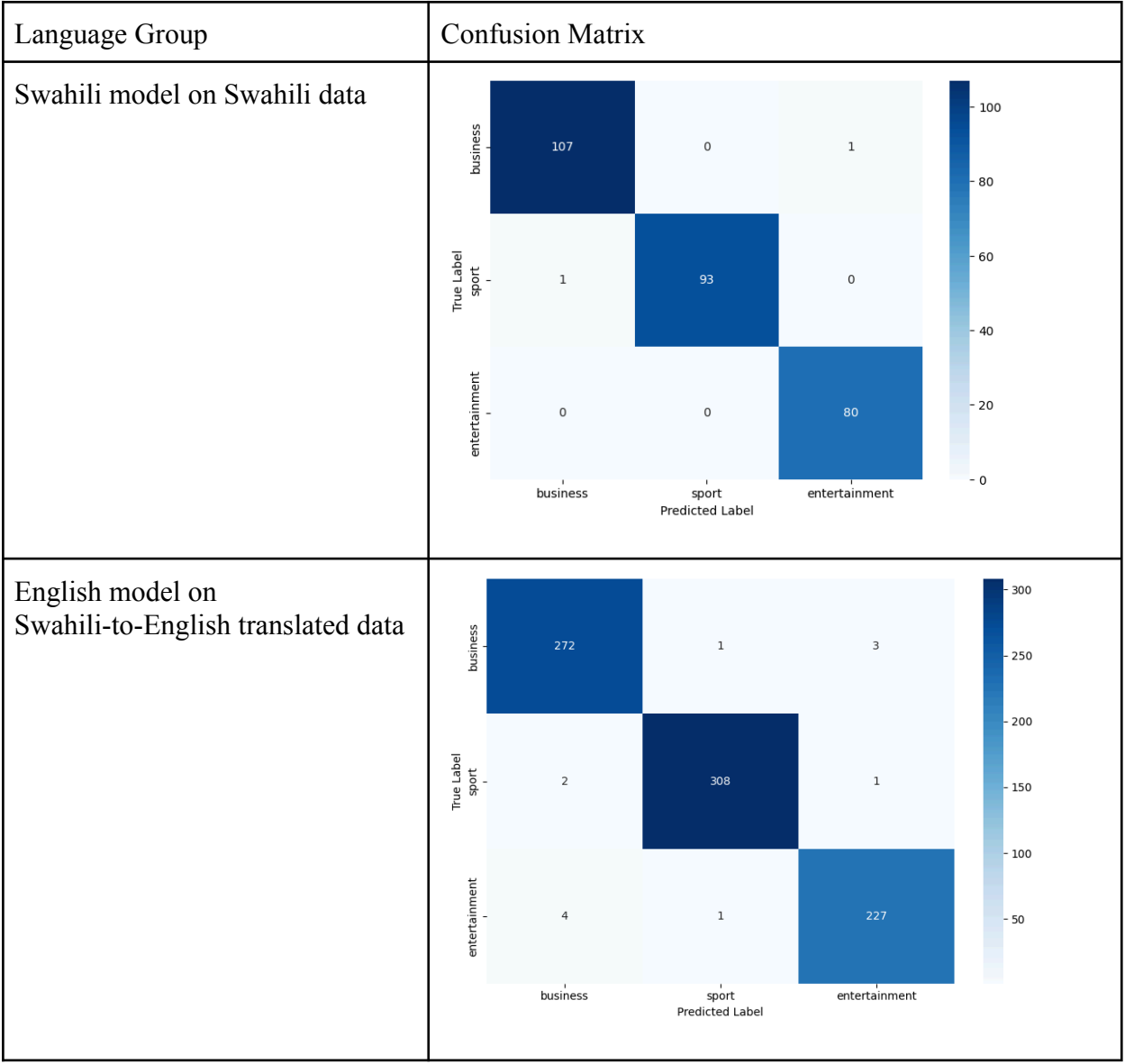 |
| English model on Swahili-to-English translated data | |

*Fig 3*. Language Group Confusion Matrices Comparison

Figure 3 demonstrates these errors through confusion matrices. The square formed where the same topic on the x- and y-axis meet (eg. business and business) is appropriately shaded a different color to highlight the number of correctly classified inputs. The numbers in the lighter boxes represent the inputs that were incorrectly classified. For example, the English model tested on Swahili-to-English translated data classified four entertainment inputs as business incorrectly (the value in the bottom-left corner).

Of course, the error rates here are small. A more difficult task for the LLM would elucidate a clearer conclusion about the nature of multilingualism and corresponding model performance. However, to the best of this study's findings, there were no datasets that existed in an appropriate form, in both English and Swahili to create BERT training environments of enhanced complexity. This further underscores the need for further research (and the accompanying data collection) into how models behave in different languages, and for equivalent datasets to exist to allow for their comparison. This would be particularly relevant for nuanced scenarios, or situations where a continuous understanding of the user's goals are required, such as in the application of language models in educational contexts.

   Some research has been done to demonstrate that students learning from language models is beneficial and when used in a multilingual context provide unprecedented access and understanding to complex materials. However, ensuring the accuracy of that information between languages is difficult, and often relies on underlying English data. A 2024 paper systematically found that when presented with a non-English token question, LLM's rely heavily on internal English tokens to think through the problem, before finally converting to the original language [14]. This conversion and abstraction is not necessarily accurate, as our experiment has indicated. Although there is some evidence that fundamental ideas within large language models are encoded with a degree of abstraction that transcends language [15] the fact that the same model does not have identical metrics across languages is evidence that this cannot totally extend all English knowledge perfectly to all other languages.

   The importance of extending such tools to communities that are under-resourced has proven to close large educational gaps between peers. Yu et al [16] found that US college environments in which under-resourced students had access to generative models, typical writing quality differences were mitigated, demonstrating that well placed LLM tools can make up for educational disparities. As other researchers have noted [17], the ability for such models to account for language, cultural and educational differences as part of their multilingual capabilities is both under researched and critical for effectively implementing Globalization 2.0 [18].

   Finally, the fact that Swahili data was more accurately understood when it was analyzed by a model that was trained exclusively in Swahili is in itself somewhat surprising. Swahili is primarily an oral language, and as the grammatical structure is fundamentally different from English, the comprehensibility of the language by BERT-like models may too be different. Despite its Latin representation, Swahili shares much more with Arabic than it does with English. In theory, this would mean that the model trained in English for this problem may have been better able to navigate the relationship between the words in each article, and correctly attribute it to one of the categories, such that the Swahili data may have benefited from that translation.

      Additionally, Swahili, like many other languages, are spoken differently based on region. For example, a person living in Dar es Salaam, Tanzania will speak a different dialect of Swahili compared to a native Nairobi, Kenya resident. These differences would also create further difficulties in the model's ability to understand the data. If the model had been built for Nairobi Swahili and was fed Dar Swahili, even more inaccuracies could arise. While such dialectical choices may also have been a factor for English, the smaller pool of data that could have been used to fine tune the original SwahBERT means that these dialectical differences may have been more pronounced.

      In reality, this experiment shows that despite differences in language structure, and that BERT was originally designed for English, it still resolves that native trained monolingual models outperform those that rely on this translation step.

## VI. Conclusion

This experiment aims to draw attention to the importance of language equity within machine learning and model development. While English remains the most widely used language on the internet, many of the communities that stand to gain the most from AI do not speak it. Additionally, the ambition to give broader access to the information and toolkits provided by AI should not be conflated with the universal acceptance of English. Truly equitable access to the many benefits of AI, such as teaching, automation and translation, should yield identical results, regardless of which language the model is queried in. This paper proves that the "pre-translation" of underrepresented language into English in order to access a capability (in this case, text classification) yields worse results. It is clear more work has to be done to allow the AI revolution to be truly global.

Significant further research is required to expose and identify exactly how best to address these challenges, and to do so in a routine and systematic way. In order to enable such research, more data should be collected on underrepresented languages, allowing for the more effective training of monolingual models (shown by this paper to be more efficient), as well as the enhanced testing and metrics for multilingual capabilities. This is critical in developing accurate multilingual knowledge representations, which is especially important in educational contexts. Ideally, a sufficiently capable multilingual model would be able to understand and process all information in all languages, with the same quality of response in all of them. However, existing research still indicates that English is the preferred language for both datasets and models. As shown by this paper, and hopefully by future research, the result of this English preference creates fundamental disparities in model performance. This may be most apparent in situations where a failure to capture nuanced or cultural meanings of words leads to model confusion, or poor knowledge representation.

Further study could be done that allows for the comparison between not just different languages, but different alphabets. When examining knowledge representation at the token level, the distinct representation of non-Latin languages makes research easier, as it is more apparent when the model is 'thinking' using them. For instance, token analysis would be more useful in determining if a model is evaluating a question in Chinese using its English knowledge, as the tokens used would be fundamentally different, compared to say English and French.

As mentioned previously, organizations and efforts such as the Mozilla Common Voice Project are accelerating the rate at which common datasets in diverse languages are being collected. However, there is still much work to be done. The reality of academic publishing is that beyond the internet, most of our repositories of knowledge are in English. Indeed, even finding data for the experiment in this project, which called for a straightforward demonstration of the issues regarding model translation and native training, were extremely limited. News classification does not scratch the surface of the possible realities of model inequalities, but without good data, it is hard to make more nuanced or technical comparisons.

Multilingual behavior exhibited by frontier LLMs is increasingly impressive. However, without proper data, continuous analysis, and engaging with diverse stakeholders, the future may inadvertently skew towards the languages, biases and cultures of the Western, English-speaking countries who develop them. If AI is to benefit all of humanity, it cannot have a bias towards those who speak a particular language.

## VII. Acknowledgements

We extend our sincerest thanks to Kathy Reid, who generously reviewed and edited this paper. As a PhD candidate at Australia National University's School of Cybernetics and Research Partner with the Mozilla Foundation's Common Voice Project, Ms. Reid brings deep expertise in voice data and dataset documentation. We are profoundly grateful for her thoughtful feedback and contributions.